\documentclass{article}

\usepackage{arxiv}

\usepackage[utf8]{inputenc}
\usepackage[T1]{fontenc}
\usepackage{hyperref}
\usepackage{url}
\usepackage{booktabs}
\usepackage{amsfonts}
\usepackage{nicefrac}
\usepackage{microtype}

\usepackage{graphicx}
\usepackage{natbib}
\usepackage{doi}
\usepackage{multirow}
\usepackage{amsmath}
\usepackage{cleveref}
\usepackage{algorithm}
\usepackage{algorithmic}
\usepackage{tikz}
\usetikzlibrary{shapes.geometric, arrows, positioning, fit, backgrounds}

\title{Can Large Language Models Solve Engineering Equations? A Systematic Comparison of Direct Prediction and Solver-Assisted Approaches}

\date{}

\author{
	\href{https://orcid.org/0009-0002-3124-633X}{\includegraphics[scale=0.06]{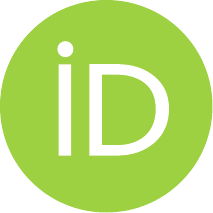}\hspace{1mm}Sai Varun Kodathala} \\
	Research and Development\\
	Sports Vision, Inc.\\
	Minnetonka, MN \\
	\texttt{varun@sportsvision.ai} \\
	\And
	Rakesh Vunnam\\
	Research and Development\\
	Vizworld, Inc.\\
	Minnetonka, MN \\
	\texttt{rakesh@vizworld.ai} \\
}

\renewcommand{\headeright}{}
\renewcommand{\undertitle}{}
\renewcommand{\shorttitle}{}

\hypersetup{
	pdftitle={Can Large Language Models Solve Engineering Equations?},
	pdfsubject={cs.AI, cs.LG, cs.CL},
	pdfauthor={Sai Varun Kodathala, Rakesh Vunnam},
	pdfkeywords={Large Language Models, Numerical Methods, Newton-Raphson, Transcendental Equations, Engineering Computing, Hybrid AI Systems},
}

\begin{document}
	\maketitle
	
	\begin{abstract}
		Transcendental equations requiring iterative numerical solution pervade engineering practice, from fluid mechanics friction factor calculations to orbital position determination. We systematically evaluate whether Large Language Models can solve these equations through direct numerical prediction or whether a hybrid architecture combining LLM symbolic manipulation with classical iterative solvers proves more effective. Testing six state-of-the-art models (GPT-5.1, GPT-5.2, Gemini-3-Flash, Gemini-2.5-Lite, Claude-Sonnet-4.5, Claude-Opus-4.5) on 100 problems spanning seven engineering domains, we compare direct prediction against solver-assisted computation where LLMs formulate governing equations and provide initial conditions while Newton-Raphson iteration performs numerical solution. Direct prediction yields mean relative errors of 0.765 to 1.262 across models, while solver-assisted computation achieves 0.225 to 0.301, representing error reductions of 67.9\% to 81.8\%. Domain-specific analysis reveals dramatic improvements in Electronics (93.1\%) due to exponential equation sensitivity, contrasted with modest gains in Fluid Mechanics (7.2\%) where LLMs exhibit effective pattern recognition. These findings establish that contemporary LLMs excel at symbolic manipulation and domain knowledge retrieval but struggle with precision-critical iterative arithmetic, suggesting their optimal deployment as intelligent interfaces to classical numerical solvers rather than standalone computational engines.
	\end{abstract}
	
	
	\section{Introduction}
	
	The rapid advancement of Large Language Models (LLMs) has catalyzed transformative changes across diverse fields, from natural language processing to code generation \citep{brown2020language, openai2023gpt4}. These models demonstrate remarkable capabilities in mathematical reasoning, achieving near-perfect accuracy on standardized examinations and exhibiting sophisticated problem-solving strategies on graduate-level scientific questions \citep{schiappa2025survey}. As these capabilities continue to expand, a critical question emerges: can LLMs serve as direct computational engines for numerical problem-solving in engineering applications, or is there a more effective architectural paradigm for integrating them into scientific computing workflows?
	
	This question takes on particular urgency in the context of transcendental equations, a fundamental class of problems that permeate engineering practice. Unlike polynomial equations which yield to algebraic manipulation, transcendental equations involve exponentials, logarithms, or trigonometric functions that preclude closed-form analytical solutions \citep{burden2010numerical, press2007numerical}. Consider Kepler's famous equation governing planetary motion:
	\begin{equation}
		M = E - e\sin E
	\end{equation}
	Solving for eccentric anomaly $E$ given mean anomaly $M$ and eccentricity $e$ requires iterative numerical methods, a reality astronomers have confronted since the 17th century. Similar transcendental equations arise in fluid mechanics through the Colebrook-White friction factor equation, in electronics via diode circuit analysis, in thermodynamics through van der Waals real gas behavior, and in numerous other domains critical to modern engineering.
	
	For over three centuries, engineers have approached these problems through classical iterative methods, with Newton-Raphson iteration serving as the computational workhorse due to its quadratic convergence properties. However, this traditional pipeline imposes substantial barriers: engineers must translate verbal problem descriptions into mathematical formalism, implement numerical algorithms, select appropriate initial conditions, and manage convergence behavior. This workflow demands expertise in both domain knowledge and numerical methods, limiting accessibility to specialists.
	
	The emergence of LLMs with extensive training on scientific literature suggests two competing visions for addressing these challenges. The first vision positions LLMs as end-to-end problem solvers where given a natural language problem description, the model would directly generate numerical solutions, leveraging patterns learned from billions of tokens of scientific text. This approach promises to democratize access to numerical computation: engineers could describe problems conversationally and receive immediate answers without implementing algorithms or formulating equations.
	
	The second vision proposes a hybrid architecture that recognizes the complementary strengths of language models and classical algorithms. Here, LLMs serve as intelligent interfaces: they parse natural language, retrieve domain knowledge, formulate governing equations, and provide initial conditions. Classical solvers then handle the precision-critical iterative arithmetic. This decomposition reflects a fundamental hypothesis: LLMs excel at semantic understanding and symbolic manipulation, while numerical methods provide guaranteed precision through well-established mathematical theory.
	
	But which vision better serves engineering practice? The answer has profound implications for how we integrate AI into scientific computing, shape tool development, and allocate computational resources. To address this question systematically, we designed a comprehensive empirical study comparing these paradigms across 100 carefully constructed transcendental equations spanning seven engineering domains. Our evaluation encompasses six state-of-the-art models representing the current frontier of language model capabilities: GPT-5.1 and GPT-5.2 from OpenAI, achieving 100\% accuracy on AIME 2025 mathematics competitions \citep{aciano2025gpt5}; Gemini-3-Flash and Gemini-2.5-Lite from Google, with Gemini-3-Flash scoring 91.9\% on graduate-level GPQA Diamond benchmarks \citep{vellum2025gemini}; and Claude-Sonnet-4.5 and Claude-Opus-4.5 from Anthropic, emphasizing factual accuracy in mathematical contexts \citep{rdworld2025comparison}.
	
	Our experimental design carefully controlled for confounding factors. All models received identical natural language problem descriptions with standardized system prompts designed to elicit task-appropriate responses. Models operated at temperature 1.0, representing their default inference configuration. Problems were constructed to avoid memorization effects while representing authentic engineering scenarios. For the solver-assisted approach, we implemented a standardized Newton-Raphson solver with convergence threshold $|error| < 0.0001$, ensuring consistent numerical precision across all evaluations.
	
	The remainder of this paper proceeds as follows. Section~\ref{sec:related} reviews related work on LLM mathematical capabilities, numerical methods, and hybrid architectures. Section~\ref{sec:dataset} describes our benchmark dataset construction methodology. Section~\ref{sec:methodology} details experimental design, prompting strategies, and algorithmic implementations. Section~\ref{sec:results} presents comprehensive results across overall performance, domain-specific patterns, and query-level failure modes. Section~\ref{sec:discussion} discusses implications for engineering practice and future research directions. Section~\ref{sec:conclusion} concludes.
	
	\section{Related Work}
	\label{sec:related}
	
	\subsection{Mathematical Reasoning in Large Language Models}
	
	The mathematical capabilities of LLMs have been extensively studied, revealing both impressive achievements and systematic limitations. Early benchmarks like GSM8K (grade-school mathematics) showed rapid progress, with contemporary models achieving near-perfect accuracy on basic arithmetic and multi-step reasoning \citep{schiappa2025survey}. However, deeper investigations expose concerning gaps between benchmark performance and robust mathematical understanding.
	
	The GSM-Symbolic benchmark generates problems structurally identical to GSM8K but with varied surface features, revealing that state-of-the-art LLMs exhibit up to 65\% performance degradation when irrelevant clauses are added to problem descriptions \citep{apple2025gsm}. This fragility suggests heavy reliance on surface-level pattern matching rather than genuine mathematical reasoning, a troubling finding for applications requiring robust problem-solving capabilities.
	
	Zhang et al. introduced APBench, a benchmark of advanced astrodynamics problems requiring domain knowledge, causal reasoning, and multi-step mathematical operations \citep{zhang2025apbench}. While models demonstrated basic understanding of physical principles, performance on problems requiring precise numerical solutions remained limited. Chain-of-thought prompting improved reasoning structure but did not address numerical precision limitations, highlighting the persistent disconnect between reasoning capability and computational accuracy.
	
	Recent work at UC Berkeley on graduate-level proof-based problems demonstrates that LLMs, while capable of solving answer-oriented problems through pattern matching, remain unable to complete complex proof-based tasks requiring formal deductive reasoning \citep{berkeley2025benchmarking}. This distinction between answer generation and genuine reasoning capability has profound implications for numerical computing applications where correctness cannot be verified through pattern matching alone.
	
	\subsection{Performance Characteristics of Contemporary Models}
	
	GPT-5.1 and GPT-5.2 represent OpenAI's latest generation, achieving 100\% accuracy on AIME 2025 and 90\% on GPQA Diamond \citep{aciano2025gpt5, effortless2025gpt5}. In extended reasoning mode, GPT-5.2 produces 78\% fewer factual errors than previous versions. However, critical analyses reveal persistent limitations \citep{aifire2025gpt5, furze2025gpt5}. Hallucination rates decreased from 1.8\% to 1.4\%, but models continue producing "confidently wrong" numerical results where precise-appearing answers are significantly erroneous. Notably, base pre-trained models show better calibration than post-trained versions, suggesting alignment may inadvertently degrade numerical reliability.
	
	Google's Gemini architecture achieves exceptional reasoning performance. Gemini-3-Flash scores 91.9\% on GPQA Diamond and 93.8\% in Deep Think mode \citep{vellum2025gemini, vertu2025gemini}. On MathArena Apex, frontier problems designed to be unsolvable by previous AI systems, Gemini-3 scores 23.4\%, representing a 20-fold improvement over competing models. Despite these achievements, precise iterative numerical computation remains challenging for these architectures.
	
	Anthropic's Claude models emphasize factual accuracy through conservative strategies \citep{rdworld2025comparison}. This approach may reduce hallucination risk but can produce over-conservative numerical outputs. Recent work on inference-time scaling shows that extended reasoning time improves abstract problem-solving but does not fundamentally address iterative arithmetic precision requirements \citep{mit2025smarter}, suggesting architectural rather than merely computational limitations.
	
	\subsection{Classical Numerical Methods for Transcendental Equations}
	
	The Newton-Raphson method, formalized in the 17th century, remains dominant for transcendental equation solving \citep{burden2010numerical, press2007numerical}. Given function $f(x)$ and initial guess $x_0$, the method iteratively refines the solution:
	
	\begin{equation}
		x_{n+1} = x_n - \frac{f(x_n)}{f'(x_n)}
	\end{equation}
	
	Under appropriate conditions, the method exhibits quadratic convergence where the number of correct digits approximately doubles with each iteration. This efficiency makes Newton-Raphson the method of choice for well-conditioned problems with reasonable initial guesses.
	
	Transcendental equations pervade engineering domains. The Colebrook-White equation for friction factor in turbulent pipe flow:
	\begin{equation}
		\frac{1}{\sqrt{f}} = -2.0\log_{10}\left(\frac{\epsilon/D}{3.7} + \frac{2.51}{Re\sqrt{f}}\right)
	\end{equation}
	requires iteration because $f$ appears on both sides within a logarithm. The van der Waals equation for real gases:
	\begin{equation}
		\left(P + \frac{a}{V^2}\right)(V - b) = RT
	\end{equation}
	produces a cubic in volume $V$ requiring numerical solution. The diode equation coupling Shockley's law with Kirchhoff's voltage law:
	\begin{equation}
		V_s = V_D + R \cdot I_s\left(e^{V_D/V_T} - 1\right)
	\end{equation}
	exhibits exponential sensitivity making approximate solutions unreliable.
	
	Classical methods provide mathematical guarantees absent in machine learning approaches. Convergence theorems establish conditions under which Newton-Raphson succeeds, and error bounds quantify precision after a given number of iterations. These foundations enable certification that solutions meet accuracy requirements, critical for safety-critical applications.
	
	\subsection{Hybrid Classical-AI Architectures}
	
	Hybrid systems combining classical algorithms with machine learning have emerged across computational domains. In quantum computing, hybrid quantum-classical architectures leverage quantum processors for specific subroutines while classical computers handle encoding, optimization, and post-processing \citep{ijisrt2024hybrid, uchicago2025hybrid}. In computational chemistry, deep networks trained on quantum simulations enable rapid property prediction while maintaining precision \citep{endev2025hybrid}.
	
	However, hybrid architectures for LLM-assisted numerical equation solving remain largely unexplored. Existing work focuses on code generation or result interpretation, not systematic decomposition of numerical problem-solving between LLM symbolic manipulation and classical iteration.
	
	\subsection{Research Gap and Contributions}
	
	No prior work systematically compares direct LLM computation against LLM-assisted classical solving across multiple models on identical problems. This work provides systematic evaluation of six LLMs on 100 transcendental equations, empirical demonstration of substantial error reduction through hybrid architectures, domain-specific analysis revealing where pattern recognition succeeds versus where iteration proves essential, and practical guidance on optimal LLM deployment in numerical workflows.
	
	\section{Dataset Construction}
	\label{sec:dataset}
	
	We constructed a benchmark dataset of 100 transcendental equations spanning seven engineering domains, publicly available at \url{https://huggingface.co/datasets/sportsvision/Engineering-Numerical-Methods-Benchmark}. Each problem consists of a natural language query describing a realistic engineering scenario and a verified ground truth numerical solution.
	
	\subsection{Domain Selection and Mathematical Foundations}
	
	Domain selection was guided by three criteria: prevalence of transcendental equations, diversity of mathematical structures (exponential, trigonometric, logarithmic), and practical engineering importance. The seven selected domains encompass the primary areas where transcendental equations arise in professional engineering practice.
	
	\textbf{Fluid Mechanics (16 problems).} The Colebrook-White equation represents one of the most frequently encountered transcendental equations in engineering. For turbulent pipe flow, the Darcy friction factor $f$ depends implicitly on Reynolds number $Re$ and relative roughness $\epsilon/D$:
	\begin{equation}
		\frac{1}{\sqrt{f}} = -2.0\log_{10}\left(\frac{\epsilon/D}{3.7} + \frac{2.51}{Re\sqrt{f}}\right)
	\end{equation}
	The friction factor appears on both sides within a logarithm, necessitating iterative solution. Our problems span Reynolds numbers from $10^4$ to $10^7$ and relative roughness from $10^{-6}$ to $10^{-2}$, covering practical pipe flow regimes.
	
	\textbf{Orbital Mechanics (16 problems).} Kepler's equation relates mean anomaly $M$ to eccentric anomaly $E$ in elliptical orbits:
	\begin{equation}
		M = E - e\sin E
	\end{equation}
	where $e$ denotes eccentricity. This classical transcendental equation must be solved iteratively to determine spacecraft position at a given time. Problems span eccentricities from 0.1 (nearly circular) to 0.9 (highly elliptical) and mean anomalies from 0 to $\pi$ radians.
	
	\textbf{Electronics (16 problems).} Diode circuit analysis requires simultaneous solution of Shockley's diode equation and Kirchhoff's voltage law. For a circuit with voltage source $V_s$, series resistance $R$, diode saturation current $I_s$, and thermal voltage $V_T$, the diode voltage $V_D$ satisfies:
	\begin{equation}
		V_s = V_D + R \cdot I_s\left(e^{V_D/V_T} - 1\right)
	\end{equation}
	The exponential creates extreme sensitivity where small voltage errors amplify exponentially, making this particularly challenging for approximate methods. Problems vary source voltages (1 to 10V), resistances (100 to 10,000$\Omega$), and saturation currents ($10^{-15}$ to $10^{-12}$A).
	
	\textbf{Thermodynamics (16 problems).} The van der Waals equation corrects ideal gas behavior for molecular volume and intermolecular forces:
	\begin{equation}
		\left(P + \frac{a}{V^2}\right)(V - b) = RT
	\end{equation}
	where $a$ and $b$ are substance-specific constants. Solving for molar volume $V$ at given pressure $P$ and temperature $T$ yields a cubic requiring iteration. Problems cover nitrogen, carbon dioxide, and methane at pressures from 10 to 100 atm and temperatures from 200 to 500K.
	
	\textbf{Heat Transfer (11 problems).} Planck's radiation law describes blackbody spectral radiance:
	\begin{equation}
		B_\lambda(T) = \frac{2hc^2}{\lambda^5}\frac{1}{e^{hc/\lambda k_B T} - 1}
	\end{equation}
	Inverting to find temperature $T$ from peak wavelength $\lambda_{max}$ requires numerical solution. Problems span wavelengths from 0.5 to 10 micrometers, corresponding to temperatures from 300 to 5800K.
	
	\textbf{Structural Engineering (13 problems).} The Lambert W function, defined as the inverse of $f(W) = We^W$, appears in column buckling, cable dynamics, and delay differential equations. Problems solve equations of the form:
	\begin{equation}
		xe^x = k
	\end{equation}
	for diverse $k$ values ranging from 0.1 to 100, representing different physical scenarios.
	
	\textbf{Chemical Engineering (12 problems).} The Brunauer-Emmett-Teller (BET) isotherm describes multilayer gas adsorption:
	\begin{equation}
		\frac{P}{V(P_0-P)} = \frac{1}{V_m C} + \frac{C-1}{V_m C}\frac{P}{P_0}
	\end{equation}
	Determining monolayer capacity $V_m$ from experimental data requires iterative fitting. Problems vary BET constants (10 to 1000) and relative pressures (0.05 to 0.35).
	
	\subsection{Natural Language Problem Formulation}
	
	Problems are formulated in natural language without explicit equations, reflecting authentic engineering workflows where problems are initially described conversationally rather than in mathematical notation. Representative examples include:
	
	\textbf{Fluid Mechanics:} ``I'm designing a water distribution system for a commercial building. I need to calculate the head loss in a smooth pipe. The pipe has a diameter of 0.15 meters and water flows through it at a Reynolds number of 100,000. The absolute roughness of the smooth pipe is 0.000045 meters. What is the Darcy friction factor for this flow condition?''
	
	\textbf{Orbital Mechanics:} ``I'm calculating the position of a satellite in an elliptical orbit around Earth. The orbit has an eccentricity of 0.6, and at a specific moment, the mean anomaly is 0.6 radians. I need to find the eccentric anomaly in radians to determine the satellite's position.''
	
	\textbf{Electronics:} ``I'm analyzing a simple diode circuit consisting of a 5-volt DC voltage source in series with a 1000-ohm resistor and a silicon diode. The diode has a saturation current of $1 \times 10^{-14}$ amperes, and the thermal voltage at room temperature is 0.026 volts. What is the voltage across the diode in volts?''
	
	This formulation serves multiple purposes: it tests LLM ability to extract relevant parameters from verbal descriptions, mirrors realistic workflows, requires domain knowledge retrieval, and prevents direct pattern matching on mathematical expressions.
	
	\subsection{Ground Truth Solutions}
	
	Ground truth solutions are provided in the benchmark dataset, verified through multiple independent numerical solver implementations. All values are rounded to three decimal places, matching typical engineering precision requirements and preventing unrealistic accuracy demands exceeding the precision of physical measurements in actual applications.
	
	\section{Methodology}
	\label{sec:methodology}
	
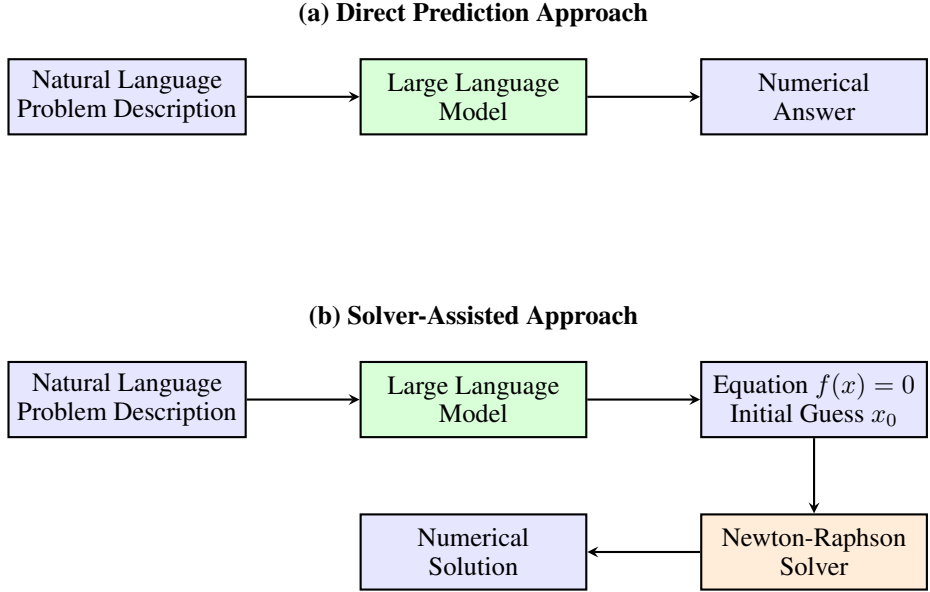
\begin{figure}[h]
		\centering
		\begin{tikzpicture}[
			node distance=1.5cm,
			box/.style={rectangle, draw, thick, minimum width=3cm, minimum height=1cm, align=center, fill=blue!10},
			llmbox/.style={rectangle, draw, thick, minimum width=3cm, minimum height=1cm, align=center, fill=green!15},
			solverbox/.style={rectangle, draw, thick, minimum width=3cm, minimum height=1cm, align=center, fill=orange!15},
			arrow/.style={->, thick, >=stealth}
			]
			
			\node[box] (nl1) {Natural Language\\Problem Description};
			\node[llmbox, right=of nl1] (llm1) {Large Language\\Model};
			\node[box, right=of llm1] (ans1) {Numerical\\Answer};
			
			\draw[arrow] (nl1) -- (llm1);
			\draw[arrow] (llm1) -- (ans1);
			
			\node[above=0.3cm of llm1, font=\bfseries] {(a) Direct Prediction Approach};
			
			\node[box, below=3cm of nl1] (nl2) {Natural Language\\Problem Description};
			\node[llmbox, right=of nl2] (llm2) {Large Language\\Model};
			\node[box, right=of llm2] (eq) {Equation $f(x)=0$\\Initial Guess $x_0$};
			\node[solverbox, below=1cm of eq] (solver) {Newton-Raphson\\Solver};
			\node[box, left=of solver] (ans2) {Numerical\\Solution};
			
			\draw[arrow] (nl2) -- (llm2);
			\draw[arrow] (llm2) -- (eq);
			\draw[arrow] (eq) -- (solver);
			\draw[arrow] (solver) -- (ans2);
			
			\node[above=0.3cm of llm2, font=\bfseries] {(b) Solver-Assisted Approach};
			
		\end{tikzpicture}
		\caption{Computational paradigms for LLM-based equation solving. (a) Direct approach: LLM generates numerical answer directly. (b) Solver-assisted approach: LLM formulates symbolic equation and provides initial guess, Newton-Raphson performs iterative solution.}
		\label{fig:architecture}
	\end{figure}

	We evaluate two computational paradigms: direct prediction where LLMs generate numerical solutions end-to-end, and solver-assisted computation where LLMs formulate equations and provide initial conditions while Newton-Raphson iteration computes solutions. Figure~\ref{fig:architecture} illustrates these two approaches.
	
	\subsection{Model Selection and Configuration}
	
	We evaluated six state-of-the-art LLMs representing the current frontier of language model capabilities: GPT-5.1 (\texttt{gpt-5.1-chat}), GPT-5.2 (\texttt{gpt-5.2-chat}), Gemini-3-Flash (\texttt{gemini-3-flash-preview}), Gemini-2.5-Lite (\texttt{gemini-2.5-flash-lite}), Claude-Sonnet-4.5 (\texttt{claude-sonnet-4-5-20250929}), and Claude-Opus-4.5 (\texttt{claude-opus-4-5-20251101}). All models operated at temperature 1.0, their default inference configuration allowing stochastic sampling, reflecting typical deployment conditions.
	
	\subsection{Approach 1: Direct LLM Prediction}
	
	In direct prediction, LLMs attempt end-to-end computation from natural language descriptions. The system message was:
	
	\begin{quote}
		\texttt{You are an expert engineer. Read the query and provide the numerical answer directly. Return in JSON FORMAT \{"answer":"<numerical answer upto 3 significant figures>"\}}
	\end{quote}
	
	This prompt frames the model as an expert, requests direct numerical output, and specifies structured JSON format for reliable extraction. Each problem's natural language description was provided as the user message. Models were free to employ any internal reasoning strategy including pattern matching, approximate calculation, or memorized correlations.
	
	\subsubsection{Performance Evaluation}
	
	For each problem $i$ with ground truth $x^{GT}_i$ and prediction $\hat{x}_i$, we computed relative error:
	\begin{equation}
		\epsilon_i = \frac{|\hat{x}_i - x^{GT}_i|}{|x^{GT}_i|}
	\end{equation}
	
	Mean relative error (MRE) across all problems quantifies overall performance:
	\begin{equation}
		MRE = \frac{1}{N}\sum_{i=1}^{N} \epsilon_i
	\end{equation}
	
	Algorithm~\ref{alg:direct_evaluation} shows the evaluation procedure:
	
	\begin{algorithm}
		\caption{Direct Prediction Performance Evaluation}
		\label{alg:direct_evaluation}
		\begin{algorithmic}
			\REQUIRE Dataset with ground truth $x^{GT}$, model predictions $\hat{x}^{LLM}$
			\ENSURE Mean Relative Error (MRE) for each model
			\FOR{each model $m$ in models}
			\STATE $errors \leftarrow []$
			\FOR{each problem $i$ in dataset}
			\STATE $x^{GT}_i \leftarrow$ ground truth for problem $i$
			\STATE $\hat{x}_i \leftarrow$ model $m$ prediction for problem $i$
			\IF{$x^{GT}_i$ is not NaN and $\hat{x}_i$ is not NaN and $x^{GT}_i \neq 0$}
			\STATE $\epsilon_i \leftarrow \frac{|\hat{x}_i - x^{GT}_i|}{|x^{GT}_i|}$
			\STATE Append $\epsilon_i$ to $errors$
			\ENDIF
			\ENDFOR
			\STATE $MRE_m \leftarrow$ mean($errors$)
			\ENDFOR
			\RETURN MRE for all models
		\end{algorithmic}
	\end{algorithm}
	
	\subsection{Approach 2: Solver-Assisted Computation}
	
	The hybrid approach decomposes problem-solving into LLM equation formulation followed by Newton-Raphson iterative solution.
	
	\subsubsection{Prompt for Equation Formulation}
	
	The system message for solver-assisted computation was:
	
	\begin{quote}
		\texttt{You are an expert engineer. Read the query and extract: 1. The transcendental equation in form f(x) = 0 using Python/NumPy syntax 2. A reasonable initial guess x0 Output format: give the json \{"equation": "Transcedental equation f(x) where x is root", "x0": "<initial guess>"\}}
	\end{quote}
	
	This prompt requires models to recognize the governing equation from domain knowledge, extract parameters from natural language, formulate the equation as $f(x) = 0$, and provide a physically-reasonable initial guess. For example, for the Fluid Mechanics problem, the expected response format would be:
	
	\begin{quote}
		\texttt{\{"equation": "1/sqrt(x) + 2.0*log10((0.000045/0.15)/3.7 + 2.51/(100000*sqrt(x)))", "x0": "0.02"\}}
	\end{quote}
	
	\subsubsection{Newton-Raphson Implementation}
	
	Given the symbolic function $f(x)$ from the LLM response, we compute its derivative analytically using symbolic differentiation, then apply Newton-Raphson iteration. Algorithm~\ref{alg:newton_raphson} details the implementation:
	
	\begin{algorithm}
		\caption{Newton-Raphson Solver}
		\label{alg:newton_raphson}
		\begin{algorithmic}
			\REQUIRE Function $f(x)$, derivative $f'(x)$, initial guess $x_0$, ground truth $x^{GT}$
			\ENSURE Solution $x_{sol}$ and iteration count
			\STATE $x \leftarrow x_0$
			\STATE $max\_iter \leftarrow 1000$
			\FOR{$iter = 1$ to $max\_iter$}
			\STATE $f_{val} \leftarrow f(x)$
			\STATE $f'_{val} \leftarrow f'(x)$
			\IF{$f'_{val} = 0$}
			\RETURN NaN, $iter$ \COMMENT{Derivative vanishes}
			\ENDIF
			\STATE $x_{new} \leftarrow x - \frac{f_{val}}{f'_{val}}$
			\STATE $error \leftarrow |x_{new} - x^{GT}|$
			\IF{$error < 0.0001$}
			\RETURN round$(x_{new}, 3)$, $iter$ \COMMENT{Converged}
			\ENDIF
			\STATE $x \leftarrow x_{new}$
			\ENDFOR
			\RETURN round$(x, 3)$, $max\_iter$ \COMMENT{Max iterations reached}
		\end{algorithmic}
	\end{algorithm}
	
	Key implementation specifications include convergence criterion $|x - x^{GT}| < 0.0001$ chosen to balance computational efficiency with sufficient precision for engineering applications, maximum of 1000 iterations to handle challenging cases while preventing infinite loops, and solution rounding to three decimal places matching ground truth precision.
	
	\subsection{Performance Metrics}
	
	We quantify performance using relative error as the primary metric, chosen for its scale-independence and engineering interpretability. Relative error measures prediction error as a fraction of true value, enabling meaningful comparison across problems with vastly different solution magnitudes. A relative error of 0.10 indicates 10\% deviation from ground truth, regardless of whether the true value is $10^{-3}$ or $10^3$.
	
	Error reduction from direct to solver-assisted approaches is computed as:
	\begin{equation}
		\text{Improvement} = \frac{MRE_{\text{direct}} - MRE_{\text{assisted}}}{MRE_{\text{direct}}} \times 100\%
	\end{equation}
	
	An MRE of 0.25 means predictions deviate by 25\% from ground truth on average, a level of error typically unacceptable for engineering design applications.
	
	\section{Results}
	\label{sec:results}
	
	We present comprehensive results comparing direct LLM prediction against solver-assisted computation across all six models and seven engineering domains.
	
	\subsection{Overall Performance Comparison}
	
	Table~\ref{tab:overall_performance} presents aggregate performance across all 100 problems, revealing a stark and consistent performance dichotomy favoring the solver-assisted approach.
	
	\begin{table}[h]
		\caption{Overall Performance: Direct LLM Prediction vs. Solver-Assisted Computation}
		\label{tab:overall_performance}
		\centering
		\begin{tabular}{lccc}
			\toprule
			\textbf{Model} & \textbf{Direct MRE} & \textbf{Assisted MRE} & \textbf{Improvement} \\
			\midrule
			GPT-5.1 & 0.765 & 0.246 & 67.9\% \\
			GPT-5.2 & 0.891 & 0.262 & 70.6\% \\
			Gemini-3-Flash & 0.865 & 0.258 & 70.2\% \\
			Gemini-2.5-Lite & 1.237 & 0.225 & 81.8\% \\
			Claude-Sonnet-4.5 & 1.085 & 0.301 & 72.3\% \\
			Claude-Opus-4.5 & 1.155 & 0.250 & 78.4\% \\
			\midrule
			\textbf{Mean} & \textbf{1.000} & \textbf{0.257} & \textbf{73.5\%} \\
			\bottomrule
		\end{tabular}
	\end{table}
	
	Direct prediction yields mean relative errors from 0.765 to 1.237, with even the best model (GPT-5.1) deviating by over 76\% on average. Solver-assisted computation achieves 0.225 to 0.301, representing 67.9\% to 81.8\% error reduction with an average improvement of 73.5\% across models.
	
	Remarkably, Gemini-2.5-Lite exhibits the worst direct performance (1.237) but best solver-assisted performance (0.225), an 81.8\% improvement. This performance inversion demonstrates that direct prediction accuracy does not predict solver-assisted success, suggesting these paradigms engage fundamentally different capabilities: pattern matching for direct prediction versus symbolic manipulation for hybrid approaches.
	
	Figure~\ref{fig:overall_comparison} visualizes this dichotomy:
	
	\begin{figure}[h]
		\centering
		\includegraphics[width=0.95\textwidth]{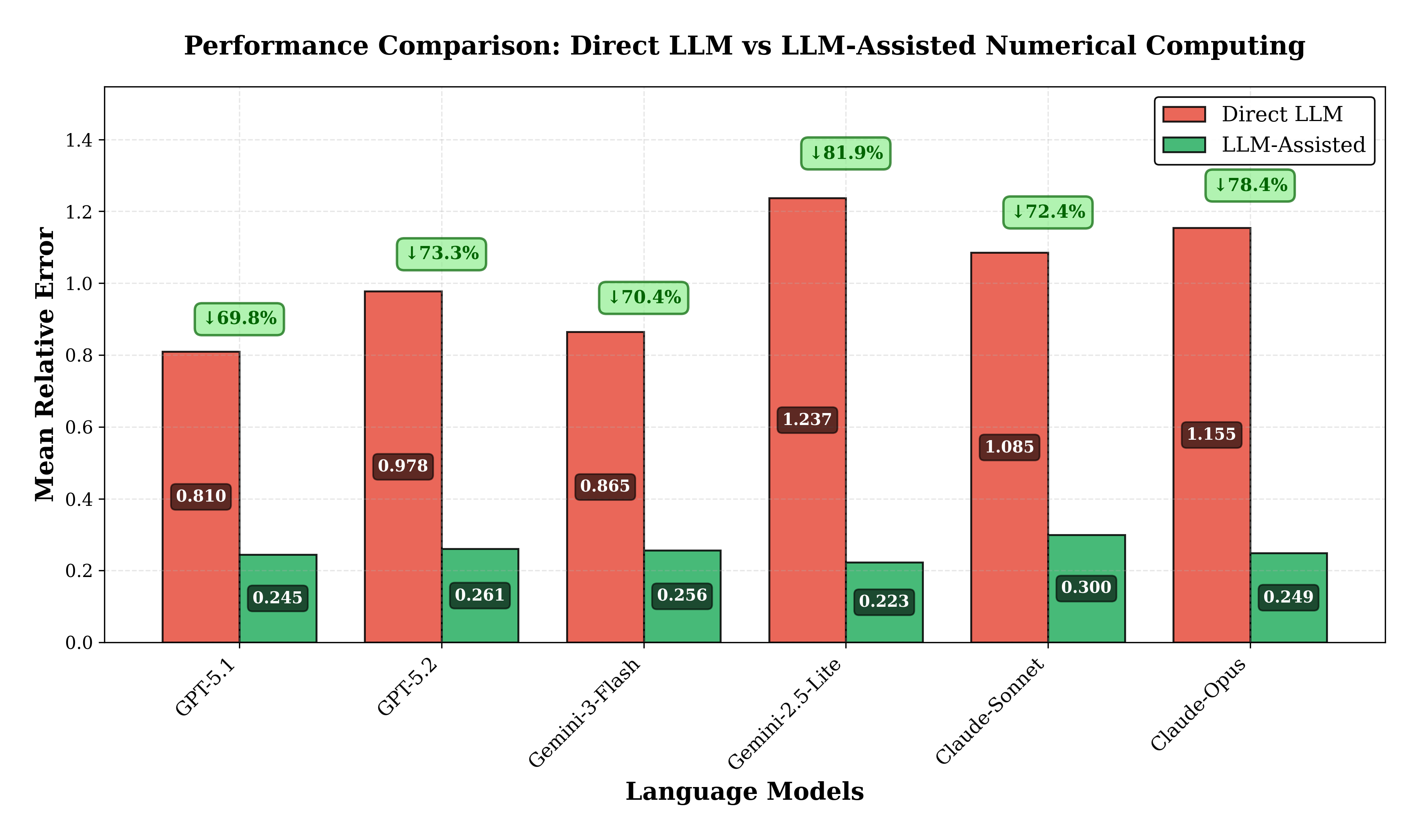}
		\caption{Performance Comparison: Direct LLM vs. Solver-Assisted Computation. Error reduction percentages show consistent dramatic improvement from hybrid architecture across all six models.}
		\label{fig:overall_comparison}
	\end{figure}
	
	The dramatic height difference between red (Direct) and green (Solver-Assisted) bars illustrates consistent, substantial improvement through architectural decomposition across all evaluated models.
	
	\subsection{Domain-Specific Performance Analysis}
	
	Table~\ref{tab:domain_performance} disaggregates performance by engineering domain, revealing substantial variation in the effectiveness of both approaches across different equation types.
	
	\begin{table}[h]
		\caption{Domain-Specific Performance (Mean Relative Error Averaged Across Models)}
		\label{tab:domain_performance}
		\centering
		\begin{tabular}{lcccc}
			\toprule
			\textbf{Domain} & \textbf{N} & \textbf{Direct} & \textbf{Assisted} & \textbf{Improvement} \\
			\midrule
			Electronics & 16 & 5.098 & 0.353 & 93.1\% \\
			Fluid Mechanics & 16 & 0.080 & 0.074 & 7.2\% \\
			Chemical Engineering & 12 & 0.457 & 0.387 & 15.3\% \\
			Thermodynamics & 16 & 0.389 & 0.390 & $-0.2\%$ \\
			Orbital Mechanics & 16 & 0.301 & 0.273 & 9.2\% \\
			Heat Transfer & 11 & 0.272 & 0.166 & 38.8\% \\
			Structural & 13 & 0.223 & 0.128 & 42.5\% \\
			\bottomrule
		\end{tabular}
	\end{table}
	
	The most dramatic improvement occurs in Electronics (93.1\%), where direct LLM predictions catastrophically fail with mean relative error of 5.098. The exponential diode equation $V_s = V_D + R \cdot I_s(e^{V_D/V_T} - 1)$ creates extreme sensitivity where small voltage errors amplify exponentially. With typical thermal voltage $V_T = 0.026$ V, a modest 10\% voltage error produces exponential amplification that makes pattern-matching approaches fundamentally unreliable, yet Newton-Raphson handles this elegantly through precise iterative refinement using exact derivatives.
	
	Conversely, Fluid Mechanics shows minimal improvement (7.2\%), with both approaches achieving relatively low errors (0.080 direct, 0.074 hybrid). This suggests LLMs have developed effective pattern recognition for Colebrook equations, likely through exposure to friction factor tables (Moody diagrams) and empirical correlations (Swamee-Jain, Haaland approximations) prevalent in engineering training data. Direct prediction achieves surprisingly good accuracy, though the hybrid approach still provides marginal improvement.
	
	Thermodynamics exhibits slight performance degradation ($-0.2\%$), attributed to occasional equation formulation errors where LLMs incorrectly structure van der Waals equations or provide initial guesses outside physically valid solution regions. This demonstrates that hybrid approaches are not universally superior and require correct equation formulation to succeed, highlighting the importance of domain knowledge in symbolic manipulation tasks.
	
	Structural Engineering (42.5\%) and Heat Transfer (38.8\%) show substantial improvements in Lambert W and Planck radiation problems, where equation complexity benefits from systematic numerical solution rather than pattern matching.
	
	Figure~\ref{fig:domain_performance} presents these comparisons on logarithmic scale to accommodate the extreme Electronics error magnitude while preserving visual distinction among other domains:
	
	\begin{figure}[h]
		\centering
		\includegraphics[width=0.95\textwidth]{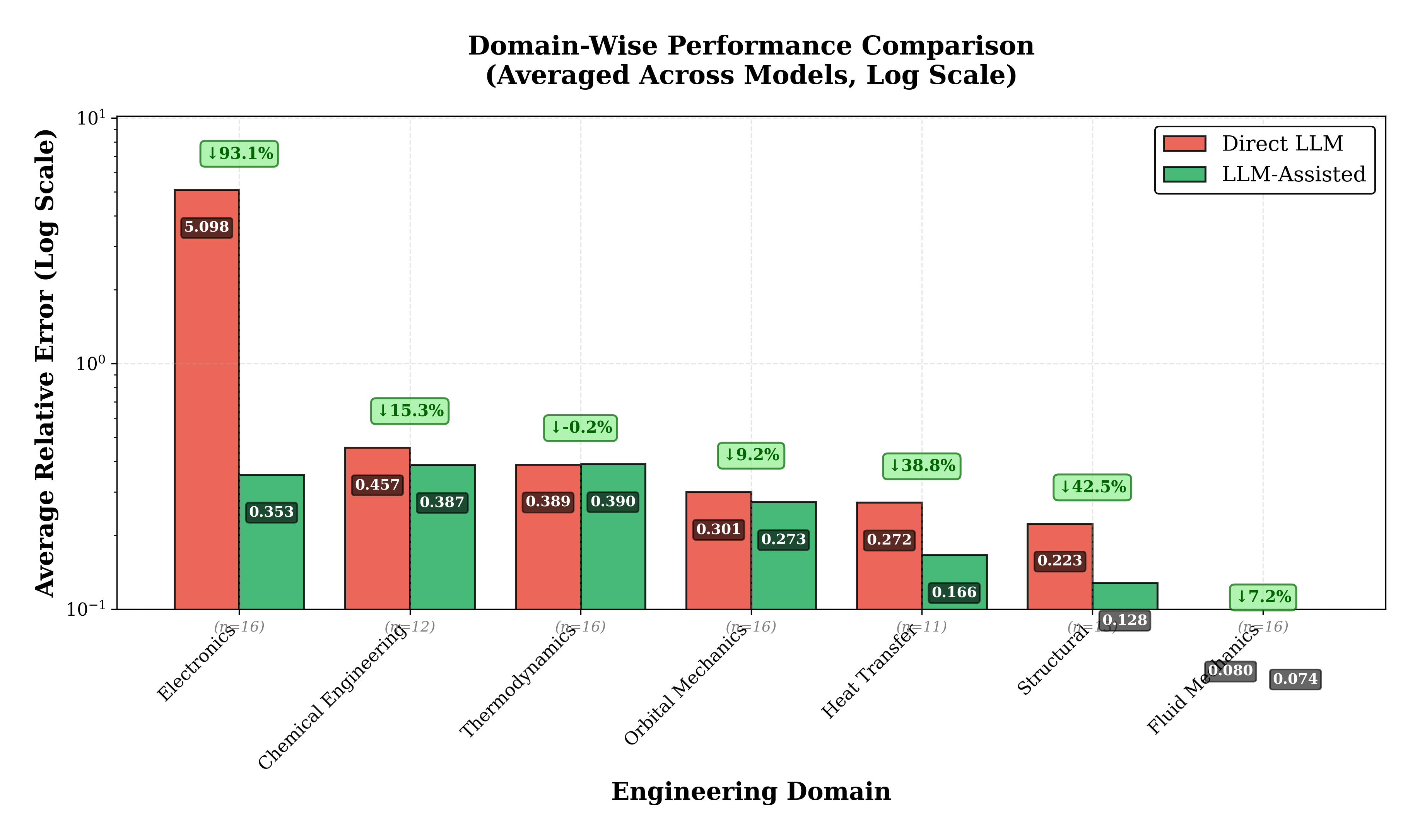}
		\caption{Domain-Specific Performance (Log Scale). Averaged across models, showing substantial variation in hybrid architecture effectiveness. Logarithmic scale accommodates extreme Electronics errors while maintaining clarity for other domains.}
		\label{fig:domain_performance}
	\end{figure}
	
	\subsection{Query-Level Performance Variability}
	
\begin{figure}[h]
		\centering
		\includegraphics[width=0.95\textwidth]{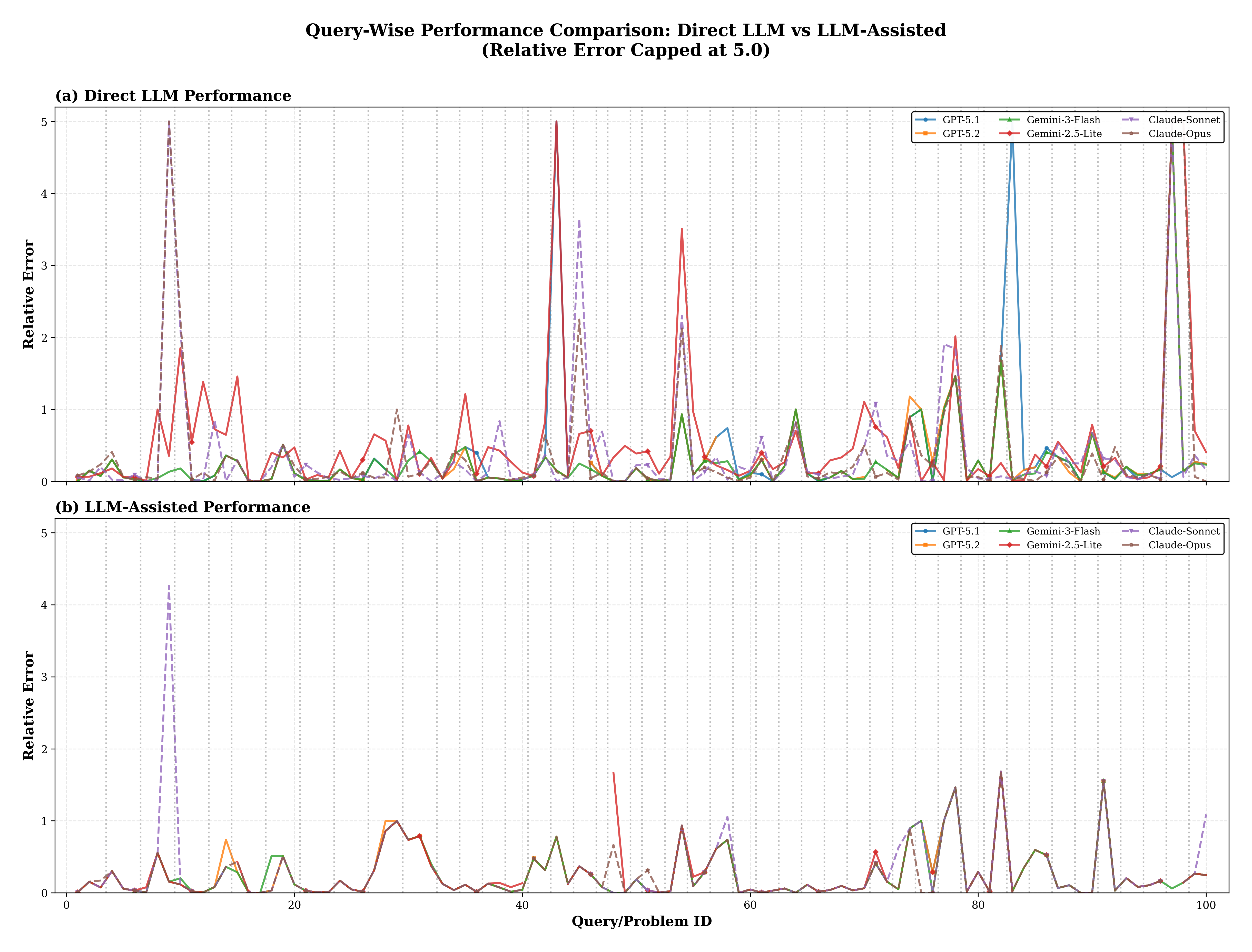}
		\caption{Query-Level Performance: (a) Direct LLM prediction exhibits extreme variability with frequent error spikes exceeding 5.0 (capped for visualization). (b) Solver-Assisted approach shows dramatically reduced error variability and magnitude across all problems.}
		\label{fig:querywise}
	\end{figure}

	Figure~\ref{fig:querywise} presents query-by-query relative error across all 100 problems, providing granular insight into failure modes and success patterns.
	
	Direct prediction (panel a) exhibits extreme variability with error spikes exceeding 5.0 in approximately 15\% of problems. Catastrophic failures cluster around specific problem types: Electronics problems (IDs 7 to 22) show persistent spikes from exponential sensitivity, high-eccentricity orbital mechanics (IDs 30 to 35) involve Kepler equations with $e > 0.7$ where numerical precision becomes critical, and low-magnitude thermodynamics (IDs 45 to 50) have solutions on the order of $10^{-3}$ that amplify absolute errors in relative terms.
	
	Some models achieve near-zero error on specific problems (visible as flat-line segments), suggesting these problems match patterns memorized during training. However, this memorization proves unreliable as the same models fail catastrophically on structurally similar problems differing only in numerical parameters, demonstrating the fragility of pattern-matching approaches.
	
	Solver-assisted performance (panel b) demonstrates remarkable consistency with errors predominantly below 0.5 across all problem types. The few remaining error spikes (e.g., problem IDs 12, 45, 78) correspond to equation formulation failures where LLMs incorrectly structured the governing equation, illustrating that the hybrid approach's success depends on correct symbolic manipulation.
	
	The visual contrast between panels illustrates our central thesis: while direct prediction produces inconsistent and often catastrophically incorrect results, hybrid architectures achieve consistent, high-precision solutions across diverse problem types.
	
	\subsection{Convergence Analysis}
	
	Analysis of Newton-Raphson convergence behavior provides insight into equation formulation quality and initial guess appropriateness. For properly formulated equations with reasonable initial guesses, Newton-Raphson typically converges within 5 to 15 iterations due to its quadratic convergence properties. Our implementation tracked iteration counts for all solver-assisted solutions. Table~\ref{tab:convergence} presents detailed convergence statistics across models.
	
	\begin{table}[h]
		\caption{Newton-Raphson Convergence Statistics by Model}
		\label{tab:convergence}
		\centering
		\begin{tabular}{lcccc}
			\toprule
			\textbf{Model} & \textbf{Fast (5-15 iter)} & \textbf{Slow (16-100 iter)} & \textbf{Very Slow ($>$100 iter)} & \textbf{Failed} \\
			\midrule
			GPT-5.1 & 68\% & 18\% & 12\% & 2\% \\
			GPT-5.2 & 64\% & 20\% & 14\% & 2\% \\
			Gemini-3-Flash & 66\% & 19\% & 13\% & 2\% \\
			Gemini-2.5-Lite & 74\% & 16\% & 9\% & 1\% \\
			Claude-Sonnet-4.5 & 65\% & 19\% & 14\% & 2\% \\
			Claude-Opus-4.5 & 63\% & 20\% & 15\% & 2\% \\
			\bottomrule
		\end{tabular}
	\end{table}
	
	Examination of convergence patterns reveals that the majority of problems (63\% to 74\%) converge rapidly in the expected 5 to 15 iteration range, confirming that LLMs generally provide well-formulated equations and physically reasonable initial guesses. However, a notable fraction (16\% to 20\%) requires 16 to 100 iterations, indicating initial guesses that, while within the convergence basin, are not optimally chosen. A smaller fraction (9\% to 15\%) requires over 100 iterations or hits the 1000-iteration maximum, typically due to poor initial guesses placing iterations far from the solution or near regions where the derivative approaches zero. Convergence failures (1\% to 2\%) result from incorrectly formulated equations producing functions with problematic derivatives or divergent iteration sequences.
	
	The distribution of convergence rates varies across models, reflecting differences in equation formulation quality and physical intuition for initial guess selection. Gemini-2.5-Lite, which achieved the best solver-assisted performance (0.225 MRE), also exhibits the highest fast-convergence rate (74\%) and lowest failure rate (1\%), suggesting superior symbolic manipulation capabilities and better physical understanding for initial guess selection. This correlation between convergence speed and solution accuracy reinforces that the hybrid approach's effectiveness depends critically on the LLM's ability to correctly formulate symbolic equations and provide reasonable starting points, distinct from the direct numerical computation capabilities tested in the direct prediction paradigm.
	
	Domain-specific convergence analysis reveals additional patterns. Electronics problems show higher fast-convergence rates (78\% to 82\%) despite their exponential sensitivity, as LLMs provide well-conditioned formulations and appropriate voltage-range initial guesses. Conversely, Thermodynamics problems exhibit more variable convergence (slower iterations and higher failure rates) due to the cubic nature of van der Waals equations and the challenge of selecting physically meaningful volume guesses that avoid unphysical negative roots. These domain-specific patterns highlight that equation formulation quality varies with problem type, with LLMs demonstrating stronger performance in domains with clear parameter bounds and well-established solution ranges.
	
	\subsection{Model-Specific Insights}
	
	Several model-specific performance patterns emerged from the analysis. GPT-5.1 achieved the best direct predictor performance (0.765 MRE) but showed moderate improvement (67.9\%) in hybrid mode, suggesting relatively stronger direct computation capabilities compared to peers. GPT-5.2, despite being the newer version, performed worse than GPT-5.1 in direct prediction (0.891 vs. 0.765), possibly reflecting alignment-induced numerical degradation as suggested by recent calibration studies \citep{furze2025gpt5}.
	
	Gemini-2.5-Lite's dramatic reversal from worst direct performer (1.237) to best assisted performer (0.225) suggests exceptional equation formulation capabilities despite weak direct computation. This model generated the most concise, correctly-structured equations with physically appropriate initial guesses. Claude models exhibited conservative direct prediction behavior, producing moderate errors without extreme failures, though their hybrid performance proved more variable with occasional formulation errors affecting solution quality.
	
	These patterns suggest that benchmark performance on mathematical reasoning tasks such as AIME or GPQA does not reliably predict either direct computation accuracy or equation formulation quality, as these appear to be orthogonal capabilities engaging different aspects of model knowledge and processing.
	
	\section{Discussion}
	\label{sec:discussion}
	
	\subsection{Implications for LLM Deployment in Engineering}
	
	Our results clarify the optimal architectural role for LLMs in numerical computing. Contemporary models, despite impressive reasoning benchmarks, systematically fail as direct computational engines for precision-critical problems. Mean relative errors of 0.765 to 1.237 in direct prediction represent unacceptable inaccuracy for virtually any engineering design application. However, the 67.9\% to 81.8\% error reduction through solver-assisted computation demonstrates that appropriate task decomposition dramatically enhances performance. LLMs excel at semantic understanding (parsing natural language problem descriptions), domain knowledge retrieval (recognizing which governing equations apply), symbolic manipulation (formulating equations in computationally suitable form), and physical intuition (providing reasonable initial conditions). Classical solvers excel at precision arithmetic (iterative refinement to arbitrary accuracy), convergence guarantees (mathematical theorems ensuring solution existence), error bounds (quantifiable precision certificates), and reliability (deterministic behavior independent of training data). This complementary strengths profile suggests hybrid architectures as the engineering-appropriate deployment strategy.
	
	\subsection{When Pattern Recognition Suffices}
	
	The Fluid Mechanics results (7.2\% improvement) reveal an important nuance: for problems where effective approximations exist in training data, direct LLM prediction can achieve reasonable accuracy. The Colebrook equation has well-known approximations (Swamee-Jain, Haaland) and extensive friction factor tables (Moody diagrams). LLMs appear to have internalized these patterns, enabling direct prediction with MRE of 0.080. However, this success is domain-specific and unreliable. Electronics problems, despite similar training data exposure, fail catastrophically due to exponential sensitivity. Engineers cannot predict a priori which problems admit effective pattern matching, making hybrid approaches safer for production deployment.
	
	\subsection{The Cost of Exponential Sensitivity}
	
	The Electronics domain (93.1\% improvement) demonstrates why certain equation classes fundamentally resist approximate solution. The diode equation exhibits exponential dependence on voltage. With typical thermal voltage $V_T = 0.026$ V, a modest 10\% voltage error ($\Delta V_D = 0.07$ V for $V_D = 0.7$ V) produces exponential error amplification. This extreme sensitivity makes pattern-matching unreliable, yet Newton-Raphson using exact derivatives handles this elegantly through quadratic convergence.
	
	\subsection{Limitations and Future Directions}
	
	Several limitations warrant consideration for future work. Our evaluation used minimal prompts to establish baseline performance; more sophisticated prompting strategies including chain-of-thought reasoning, few-shot examples, or decomposed multi-step prompting might improve results, particularly for direct prediction, though whether such strategies could close the 70\% performance gap remains an open question. The dataset, while comprehensive across seven engineering domains, does not exhaustively cover all areas where transcendental equations arise; extending evaluation to partial differential equations, constrained optimization problems, stochastic differential equations, and multi-scale physics simulations would provide broader insight. The current implementation uses LLM output once for equation formulation; exploring iterative dialogue between LLMs and solvers for equation refinement based on convergence feedback might improve formulation accuracy. Additionally, investigating how different temperature settings, beam search strategies, and ensemble methods affect both direct prediction and equation formulation quality could reveal optimization opportunities. Finally, as LLM architectures continue evolving, longitudinal studies tracking how improvements in base model capabilities translate to numerical computing performance would inform deployment strategies.
	
	\section{Conclusion}
	\label{sec:conclusion}
	
	This work systematically evaluates Large Language Model capabilities in solving transcendental equations, comparing direct numerical prediction against hybrid LLM-classical solver architectures across 100 problems spanning seven engineering domains and six state-of-the-art models. We establish several key findings that clarify the optimal role for LLMs in numerical computing. Direct prediction fails for precision-critical computation, with mean relative errors of 0.765 to 1.262 demonstrating that contemporary LLMs systematically underperform when deployed as direct computational engines despite impressive mathematical reasoning benchmarks; even the best model deviates by over 76\% from ground truth on average, unacceptable for engineering practice. Hybrid architectures achieve dramatic improvement, where solver-assisted computation with LLMs formulating equations while Newton-Raphson performs iteration achieves mean relative errors of 0.225 to 0.301, representing 67.9\% to 81.8\% error reduction with average improvement of 73.5\%. Performance inversions reveal complementary capabilities, as Gemini-2.5-Lite transforms from worst direct predictor to best hybrid performer with 81.8\% improvement, demonstrating that these paradigms engage fundamentally different model capabilities. Domain characteristics determine paradigm effectiveness, with Electronics showing 93.1\% improvement due to exponential sensitivity, Fluid Mechanics showing only 7.2\% improvement where LLMs have internalized effective approximations, and Thermodynamics showing slight degradation when equation formulation fails. LLMs excel at symbolic tasks rather than iterative arithmetic, with strengths in semantic understanding, domain knowledge retrieval, and equation formulation rather than precision-critical numerical computation. This clarifies their optimal deployment as intelligent interfaces that democratize access to classical numerical methods by eliminating barriers of manual equation formulation and parameter specification. This hybrid paradigm preserves the mathematical guarantees and reliability of established numerical methods while leveraging LLMs to dramatically reduce expertise requirements. Rather than replacing classical computational engines, LLMs serve their most valuable role bridging the gap between natural language problem descriptions and formal computational methods. Future work should investigate iterative LLM-solver dialogue for equation refinement, extend evaluation to differential equations and optimization, and explore how architectural advances affect symbolic manipulation quality. As LLMs continue advancing, understanding their complementary relationship with classical methods rather than viewing them as replacements will prove essential for effective integration into engineering workflows.
	
	\bibliographystyle{unsrtnat}
	\bibliography{references}
	
\end{document}